\documentclass{article}

\usepackage[main, final]{neurips_2026}

\usepackage[utf8]{inputenc} 
\usepackage[T1]{fontenc}    
\usepackage{hyperref}       
\usepackage{url}            
\usepackage{booktabs}       
\usepackage{amsfonts}       
\usepackage{nicefrac}       
\usepackage{microtype}      
\usepackage{xcolor}         
\usepackage{amsmath}
\usepackage{graphicx}

\title{Humanoid Horizon: Extending Task Horizon in Whole-Body Loco-Manipulation via Parallel Training, Dynamic Starting, and Reward Gating}

\author{%
  Haozhuo Zhang \\
  The University of Manchester
  \And
  Qiang Zhang \\
  University of Science and \\
  Technology of China
  \And
  Jian Tang \\
  X-Humanoid
  \AND
  Mingzhe Ni \\
  The University of Manchester
  \And
  Michele Caprio \\
  University of Warwick
  \AND
  Angelo Cangelosi\thanks{Corresponding authors.} \\
  The University of Manchester
  \And
  Wei Pan\footnotemark[1] \\
  Newcastle University
}

\begin{document}

\maketitle

\begin{abstract}
Cluttered indoor environments, where large and heavy objects are scattered across diverse surfaces, require humanoid robots to sequentially navigate, grasp, transport, and accurately place each item at its target location within a single uninterrupted episode. This long-horizon, whole-body loco-manipulation task remains a significant challenge for current methods. Previous approaches often suffer from two main issues: easy-reward bias, where training overemphasizes early transport stages at the expense of later ones, and catastrophic forgetting, where focusing on later stages leads to a decline in earlier-stage performance. In this work, we introduce Humanoid Horizon, a unified policy framework designed to overcome these limitations through three interrelated mechanisms. The Parallel Training Strategy organizes $N$ scenes into $S$ concurrent stage streams governed by a shared policy, ensuring all transport stages receive continuous gradient updates and removing the bottleneck of sequential optimization. The Dynamic Starting Mechanism updates each environment's initial state with terminal states from upstream rollouts, gradually broadening transition coverage and enhancing robustness at stage boundaries. Reward Gating sets the reward to zero for the rest of the episode in later-stage streams when the immediately preceding object is displaced beyond a set threshold, so the shared policy learns not to disturb a just-placed object and earlier placements are preserved throughout the episode. Collectively, these strategies achieve per-stage success rates exceeding 80\% on the two-object LHM-Humanoid benchmark (350 training scenes, 66 held-out scenes). As the number of sequentially transported objects grows beyond two, success declines with the horizon, but the degradation is graceful relative to the sharp drop seen in all baselines. Additionally, we demonstrate that the RL teacher policy can be distilled into a Vision-Language-Action student via DAgger. When provided only with egocentric RGB or depth observations and natural language instructions, the student successfully replicates the teacher's long-horizon, multi-object behaviors, highlighting the potential for perception-driven deployment on real humanoid robots. Videos and results are available at \url{https://haozhuo-zhang.github.io/Humanoid-Horizon-project-page/}.
\end{abstract}

\section{Introduction}

Humanoid agents that imitate human motion, perceive cluttered scenes, and manipulate objects are a central yet difficult research target. Prior work has advanced locomotion and control~\cite{peng2021amp, tessler2023calm, luo2023perpetual, zhu2023neural}, scene interaction~\cite{starke2019neural, hassan2023synthesizing, xiao2023unified, pan2024synthesizing}, and manipulation~\cite{xu2024humanvla, wang2023physhoi, xie2023hierarchical, lin2025simgenhoiphysicallyrealisticwholebody}. A key gap remains: long-horizon whole-body loco-manipulation, i.e., continuously transporting multiple large objects over long task chains without environment resets.

This gap is fundamental. Compared with arms or quadrupeds, humanoids are less dynamically stable and must coordinate balance, locomotion, and manipulation throughout long episodes. As a result, long-horizon pipelines developed for more stable platforms are hard to transfer directly. Despite progress in interface and perception conditioning~\cite{xu2024humanvla, ding2025humanoid}, most current methods still focus on short-horizon single-object skills~\cite{wang2024skillmimic, xu2025intermimic, pan2025tokenhsi}, often relying on fixed starts, and thus struggle with continuous multi-object loco-manipulation. When these methods are extended to longer task chains, two failure modes become prominent: easy-reward bias (over-optimizing early-stage transport while under-exploring later-stage behaviors) and catastrophic forgetting. In our context, catastrophic forgetting specifically means that a policy already capable of single-object transport is further trained from post-first-object terminal states to transport the next object, yet after this shift it loses the ability to re-execute first-object transport-and-release, even under substantial robot-initial-state randomization. LHM-Humanoid~\cite{zhang2026lhmhumanoidlearningunifiedpolicy} provides a concrete instance of this behavior (e.g., teacher model~2), which then motivates dual-teacher distillation into a student and, in turn, constrains scalability to longer multi-object horizons.

\begin{figure}[!t]
  \centering
  \includegraphics[width=1.0\linewidth]{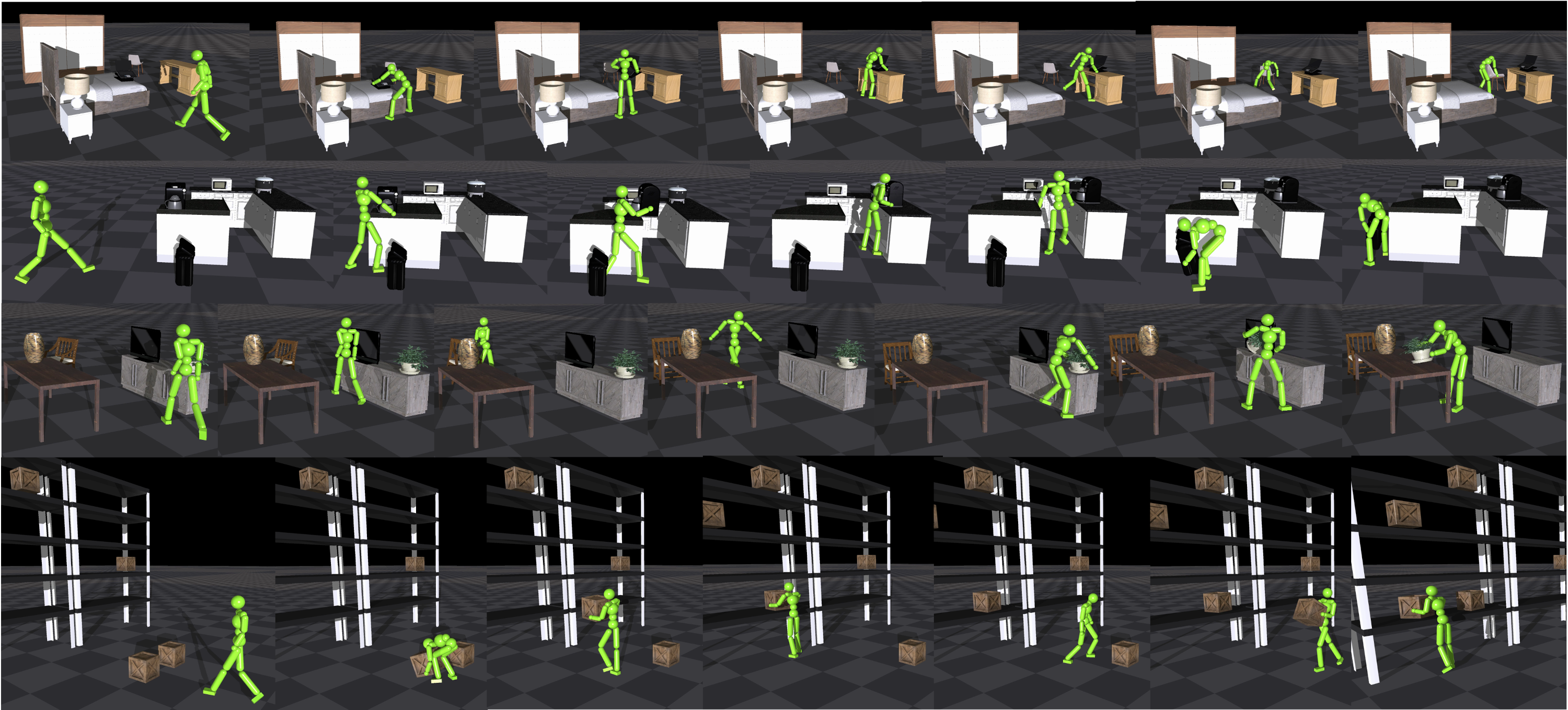}
  \caption{Humanoid Horizon uses a single policy for long-horizon whole-body loco-manipulation across cluttered bedroom, kitchen, living room, and warehouse scenes without resets. Each row shows one uninterrupted multi-object transport episode with sequential navigation, grasping, carrying, and placement.}
  \label{fig:first-img}
\end{figure}

We present Humanoid Horizon, a framework for long-horizon whole-body humanoid loco-manipulation that uses one policy to perform continuous long-horizon loco-manipulation across multiple objects, without resets, auxiliary models, or handcrafted task-specific modules; it has three coupled components: Parallel Training Strategy, which jointly optimizes all transport stages in one policy to avoid easy-reward bias in sequential curricula and forgetting risks in decoupled stage-wise training (including distillation-heavy pipelines such as LHM-Humanoid)~\cite{zhang2026lhmhumanoidlearningunifiedpolicy}, enabling scalable learning as object count grows; Dynamic Starting Mechanism, which writes each stage terminal state to the corresponding downstream environment buffer and initializes the next stage from that buffer in the next epoch, so coverage of transition states expands through temporal accumulation across epochs; and Reward Gating, which zeros the remaining reward in a later stage once the immediately preceding object is displaced beyond a tolerance, so the policy learns not to disturb a just-placed object and earlier placements are preserved over the episode.

We evaluate on the LHM-Humanoid benchmark~\cite{zhang2026lhmhumanoidlearningunifiedpolicy}: 350 training/evaluation scenes and 66 unseen scenes for generalization, and we follow LHM-Humanoid to test longer-horizon settings with more than two sequentially transported objects. Combining these mechanisms yields per-stage success above 80\% in two-object sequential transport. As more objects are carried in one episode, success declines with the horizon, but the degradation is graceful relative to the sharp drop seen in prior baselines, and a single policy still completes multi-object rearrangement across diverse scenes. As an exploratory deployment study, we distill the teacher policy into a vision-language-action (VLA) student using DAgger~\cite{ross2011reduction}, and with egocentric RGB or depth inputs plus language instructions, the student preserves long-horizon multi-object behaviors learned by the teacher.

Our contributions are summarized as follows:
\vspace{-0.5em}
\begin{itemize}
  \item We present \textbf{Humanoid Horizon}, a single-policy framework for long-horizon whole-body humanoid loco-manipulation that supports continuous multi-object rearrangement without resets, auxiliary models, or handcrafted task modules.
  \item We introduce a \textbf{Parallel Training Strategy} that jointly optimizes all transport stages, reducing easy-reward bias and catastrophic forgetting, and improving scalability to longer task horizons.
  \item We propose a \textbf{Dynamic Starting Mechanism} and \textbf{Reward Gating} that together improve inter-stage transition robustness and preserve previously completed placements throughout long-horizon execution.
  \item We demonstrate strong long-horizon performance on 350 benchmark scenes plus 66 unseen scenes, and show that DAgger-based VLA distillation transfers the learned behavior to egocentric RGB/depth inputs.
\end{itemize}

\section{RELATED WORK}

\subsection{Physically-Based Motion Synthesis and Human-Scene Interaction}

Motion synthesis spans computer graphics, vision, and robotics, and is commonly divided into kinematics-based methods\cite{jiang2024scaling, li2023object} and physics-based methods\cite{tessler2023calm, zhu2023neural, peng2021amp, peng2018deepmimic}. Kinematic approaches emphasize semantic coherence, while physics-based approaches learn dynamically feasible control policies in simulation. Foundational systems such as AMP\cite{peng2021amp} and DeepMimic\cite{peng2018deepmimic} have enabled later extensions with adversarial priors, latent skill representations, and language conditioning. In human-scene interaction, InterScene\cite{pan2024synthesizing} and InterPhys\cite{hassan2023synthesizing} move beyond earlier inverse-kinematics or hand-designed controllers toward dynamic object interaction, and RL-based methods show strong manipulation skills\cite{xu2025intermimic, wang2024skillmimic}. In parallel, embodied AI studies integrated navigation and manipulation\cite{yenamandra2023homerobot, weihs2021visual}, including commonsense-guided policies\cite{wu2023tidybot}. However, most systems still use simplified embodiments (e.g., mobile bases with parallel grippers) and remain short-horizon and single-object. HumanVLA\cite{xu2024humanvla} advances humanoid manipulation of heavier objects but remains limited to single-object tasks with fixed initial states. LHM-Humanoid\cite{zhang2026lhmhumanoidlearningunifiedpolicy} is an important step toward continuous multi-object vision-language loco-manipulation, yet its sequential training and task-specific designs limit scalability: as horizons grow, catastrophic forgetting and easy-reward bias become pronounced. Our method addresses these issues with a unified policy using Parallel Training Strategy,  Dynamic Starting Mechanism, and Reward Gating, enabling robust, scalable long-horizon multi-object loco-manipulation without environment resets.

\subsection{Long-Horizon Loco-Manipulation}

Long-horizon execution has been studied in fixed-base arms~\cite{wang2023mimicplay,lin2024hierarchical,zhao2024tac}, mobile manipulation~\cite{jiao2021efficient,zhi2025closed,zhi2025learning}, and aerial manipulation, but remains difficult because long action chains accumulate physical constraints~\cite{bu2025univla,fan2025diffusion,feng2025reflective,zhang2024navid}. Typical strategies decompose tasks via subgoals~\cite{shridhar2023perceiver,borja2022affordance}, world-state prediction~\cite{feng2025reflective}, trajectory prediction~\cite{fan2025diffusion}, or historical observations~\cite{wang2025lola}, yet often depend on task-specific priors and generalize poorly. Progress is more limited in humanoids: most teleoperation systems remain short-horizon and open-loop~\cite{he2024omnih2o,cheng2024expressive,ji2024exbody2} because reliable real-time bipedal state estimation is hard, and although legged odometry has advanced~\cite{wisth2022vilens,ou2024leg,allshire2025visual}, transfer to humanoids is still limited (CLONE~\cite{li2025clone} only partially mitigates drift with LiDAR odometry). Autonomous long-horizon whole-body humanoid loco-manipulation is therefore still nascent, with representative attempts including LessMimic~\cite{lin2026lessmimic} and LHM-Humanoid~\cite{zhang2026lhmhumanoidlearningunifiedpolicy}. As horizons grow, these methods still face the same two bottlenecks discussed above: catastrophic forgetting and easy-reward bias. Motivated by this gap, Humanoid Horizon targets horizon scalability by maintaining stage-wise competence while preserving previously completed placements as task length increases. Learning those stage transitions also depends on where each stage begins. Reverse-curriculum learning~\cite{florensa2017reverse} grows the set of start states backward from the goal, and so needs the agent to be reset into hand-specified near-goal configurations. Reset-free autonomous RL~\cite{eysenbach2017leave} avoids manual resets by training a separate reset or backward policy, with its own reward. Curricula taken from expert demonstrations~\cite{dai2021automatic} start the agent from demonstrated states. Dynamic Starting (Sec.~\ref{subsec:dynamic_starting}) is aimed at the handoff between stages instead: the next stage begins from the terminal state the preceding stage has just produced, with no reset policy, demonstration source, or extra reward.

\section{Method}

This section presents the Humanoid Horizon training pipeline with a focus on the two central goals introduced earlier: extending effective task horizon and mitigating forgetting in long sequential whole-body manipulation. We begin with the task formulation, then present our Parallel Training Strategy, which re-structures temporally sequential object transport into a unified optimization problem that preserves earlier-stage competence while learning later stages. We next describe the Dynamic Starting Mechanism, which improves stage-to-stage continuity by initializing each stage from naturally occurring intermediate states rather than fixed resets. We then introduce the Reward Gating mechanism to ensure that later-stage transport does not disrupt objects placed in earlier stages. Finally, as an extension, we introduce our distillation pipeline that transfers the reinforcement learning-trained teacher into a vision-language-action (VLA) student policy operating solely from egocentric RGB/depth observations and natural language instructions.

\subsection{Task Formulation}
\label{subsec:task_formulation}

We define continuous multi-object rearrangement over a set of scenes $\mathcal{N}=\{1,2,\ldots,N\}$. Each scene $n\in\mathcal{N}$ has its own layout, object configuration, and task specification, and contains an ordered object set $\mathcal{O}_n=\{o_1^n,o_2^n,\ldots,o_{S_n}^n\}$. The humanoid must move each object from initial positions $\{p_{\text{init},1}^n,\ldots,p_{\text{init},S_n}^n\}$ to target positions $\{p_{\text{goal},1}^n,\ldots,p_{\text{goal},S_n}^n\}$. For each object $o_i^n$, execution has four stages: (1) navigate to the object, (2) grasp and lift, (3) transport toward the goal, and (4) place at the designated target. After finishing $o_i^n$, the humanoid transitions immediately to $o_{i+1}^n$ without environment reset, so the policy must handle turning, obstacle avoidance, and re-navigation under evolving intermediate states. The benchmark spans diverse placements and semantics: objects and goals may lie on shelves, beds, floors, tables, cabinets, desks, or designated floor regions, and object categories include bedside tables, laptops, storage boxes, coffee machines, and vases. Humanoid Horizon aims to learn a single policy $\pi_{\theta}:\mathcal{X}\rightarrow\mathcal{A}$ that generalizes across all $N$ scenes and completes each full sequence $\{o_1^n,o_2^n,\ldots,o_{S_n}^n\}$ under one control framework, without policy switching or external intervention. Here, $n$ indexes scenes ($n\in\mathcal{N}$), while $\mathcal{X}$ and $\mathcal{A}$ denote the shared robot--environment state space and whole-body control action space, respectively.

\begin{figure}[!t]
  \centering
  \includegraphics[width=1.0\linewidth]{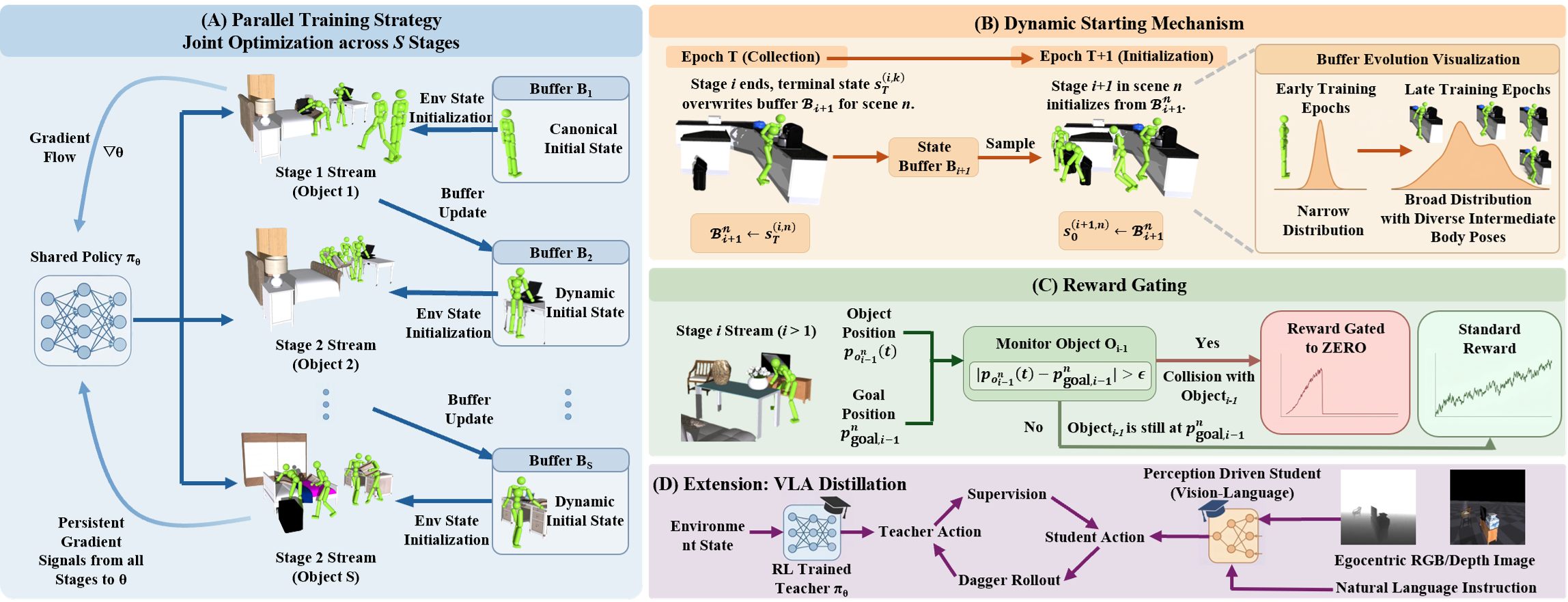}
  \caption{Overview of the Humanoid Horizon training pipeline. (A) \textbf{Parallel Training}: $S$ stage streams share one policy $\pi_\theta$ and are optimized jointly, each initialized from its state buffer ($\mathcal{B}_1$--$\mathcal{B}_S$). (B) \textbf{Dynamic Starting}: for each scene $n$, the terminal state from stage~$i$ overwrites the corresponding downstream buffer $\mathcal{B}_{i+1}^n$ (one state per environment) to initialize stage~$i{+}1$ in the next epoch; repeated overwrites across epochs broaden transition-state coverage over time. (C) \textbf{Reward Gating}: for stage-$i$ streams ($i>1$), reward is zeroed if object $o_{i-1}$ moves beyond tolerance $\epsilon$, preserving previously completed placements. (D) \textbf{VLA Distillation}: the RL teacher is distilled via DAgger into a vision-language-action student conditioned on egocentric RGB/depth observations and language instructions.}
  \label{fig:main-img}
\end{figure}

\subsection{Parallel Training Strategy}
\label{subsec:parallel_training}

Prior long-horizon humanoid rearrangement methods---including end-to-end RL, curriculum RL, and stage-isolated sequential training---are constrained by serial optimization, which tends to induce easy-reward bias toward early transport stages or catastrophic forgetting when later stages are added. To remove this bottleneck, we adopt a Parallel Training Strategy that jointly optimizes all transport stages within one policy. We first pretrain on single-object transport to establish basic loco-manipulation competence (navigate, grasp/lift, and carry/place from fixed starts), then switch to parallel training over $N$ scenes with $S$ sequential objects each (Sec.~\ref{subsec:task_formulation}). These scenes are organized into $S$ stage streams that share parameters $\theta$. One pool of parallel environments is divided evenly across the streams, and the baselines use the same pool (Appendix~\ref{app:rl_hyperparams}).

For stream $i$, training starts from buffer $\mathcal{B}_i$: $\mathcal{B}_1$ uses the original initial-state distribution, while $\mathcal{B}_i$ ($i>1$) approximates post-stage-$(i{-}1)$ intermediate states. Initially, each environment buffer in $\mathcal{B}_i$ ($i>1$) is set to a simple proxy state (robot near object-$(i{-}1)$ goal with default pose), and these coarse starts are later replaced by rollout-derived states via the Dynamic Starting mechanism (Sec.~\ref{subsec:dynamic_starting}). Keeping all $S$ streams active throughout training provides persistent gradients for every stage, suppressing easy-reward bias and forgetting; importantly, stage-$i$ streams ($i>1$) also force learning of transition-specific behaviors absent in stage~1 (release, reorientation, obstacle-aware maneuvering, and walking from non-canonical poses). Overall, this strategy yields stable optimization for long-horizon continuous rearrangement under a single unified policy.

\subsection{Dynamic Starting Mechanism}
\label{subsec:dynamic_starting}

While Parallel Training addresses easy-reward bias and catastrophic forgetting, it does not by itself solve stage-to-stage transition robustness in long-horizon execution. As described in Sec.~\ref{subsec:parallel_training}, each stage-$i$ stream ($i>1$) starts from a fixed proxy buffer $\mathcal{B}_i$; this is sufficient for bootstrapping isolated transport skills, but it creates a distribution mismatch because fixed proxies represent only a narrow, idealized subset of post-stage-$(i{-}1)$ states. The mismatch worsens with horizon length: even successful rollouts end in diverse handoff configurations (orientation, foot placement, residual momentum, and goal-relative offset), so stage-$(i{+}1)$ policies trained from fixed starts become brittle when preceding stages terminate in unfamiliar states.

We therefore introduce Dynamic Starting, which progressively replaces proxy starts with rollout-derived terminal states. At the end of each epoch, for each scene $n$, we overwrite the corresponding downstream stage buffer (which stores one state per environment) with the stage-$i$ terminal state:
\begin{equation}
  \mathcal{B}_{i+1}^{n} \leftarrow s_T^{(i,n)},
  \label{eq:ds_write}
\end{equation}
where $s_T^{(i,n)}$ is the terminal state of scene $n$ in the stage-$i$ rollout of the current epoch. In the next epoch, scene $n$ in stage-$(i{+}1)$ is initialized by directly reading this updated buffer:
\begin{equation}
  s_0^{(i+1,n)} \leftarrow \mathcal{B}_{i+1}^{n}.
  \label{eq:ds_read}
\end{equation}

Although each scene-stage buffer stores only one state at any instant, repeated epoch-wise overwrites make the policy train on a temporally accumulated sequence of handoff states. As training improves, these recorded terminal states become both more competent and more diverse over time, creating a positive feedback loop: better stage-$i$ behavior yields better downstream starts, which improves stage-$(i{+}1)$ robustness and subsequently its own terminal-state quality. Compared with fixed proxy initialization, Dynamic Starting better covers natural inter-stage transitions without extra supervision, handcrafted sampling, or scene-specific trajectories. It is complementary to Parallel Training: Parallel Training supplies persistent gradients to all stages, while Dynamic Starting aligns each stage's initial-state sequence with the true handoff dynamics from the preceding stage. Together, they improve stability, robustness, and scalability in long-horizon continuous loco-manipulation.

\subsection{Reward Gating for Physical Consistency}
A unique challenge in continuous multi-object rearrangement is that success requires every transported object to remain at its goal throughout the entire episode. We define the Success-All criterion: an episode succeeds only if all $S$ objects are placed at their respective goals and remain undisturbed, contrasting with per-stage metrics that ignore collateral disturbance of previously placed objects.

When transitioning from transporting object $o_{i-1}^n$ to object $o_i^n$, the humanoid must release, reorient, and navigate within a cluttered environment where already-placed objects may obstruct the path. Without explicit supervision, the policy may inadvertently displace previous objects during this transition, violating Success-All even if each individual transport succeeds.

To address this, we introduce Reward Gating for stage-$i$ streams with $i > 1$. While all streams share the same reward function, the stage-$i$ stream monitors object $o_{i-1}^n$ throughout rollout. If the robot displaces $o_{i-1}^n$ from its goal (by collision, pushing, or tipping), reward is immediately zeroed for the remainder of the episode:
\begin{equation}
  r_t^{(i)} \leftarrow 0 \quad \text{if } \|p_{o_{i-1}^n}(t) - p_{\text{goal}, i-1}^n\| > \epsilon,
\end{equation}
where $p_{o_{i-1}^n}(t)$ denotes the position of object $o_{i-1}^n$ at time $t$ and $\epsilon$ is a tolerance threshold. Stage $i>1$ therefore gates only on $o_{i-1}^n$. The stages share one policy $\pi_\theta$, which learns a single index-independent behavior: executing stage $i$ keeps
\begin{equation}
  \|p_{o_{i-1}^n}(t) - p_{\text{goal}, i-1}^n\| \le \epsilon.
  \label{eq:rg_local}
\end{equation}
At test time the same policy runs stages \(i=2,\ldots,S\) in order. Substituting \(i=k+1\) at each boundary yields
\begin{equation}
  \|p_{o_k^n} - p_{\text{goal}, k}^n\| \le \epsilon, \qquad k=1,\ldots,S-1.
  \label{eq:rg_chain}
\end{equation}
Eq.~\eqref{eq:rg_chain} is Eq.~\eqref{eq:rg_local} applied once per stage. Each object is guarded exactly when it is the immediately preceding one, and preservation of \(\{o_1^n,\ldots,o_{S-1}^n\}\) emerges from that recursion alone.

\subsection{Extension: Vision-Language-Action Policy Distillation}
\label{subsec:vla_distillation}

As an exploratory step toward real-world deployment, we distill the RL-trained teacher policy into a Vision-Language-Action (VLA) student model using DAgger~\cite{ross2011reduction}, which iteratively collects on-policy student rollouts and queries the teacher for corrective supervision, mitigating the compounding distribution shift that would otherwise arise from behavior cloning over long horizons. We train two student variants: one conditioned on egocentric RGB images and one on egocentric depth maps, both conditioned on per-stage natural language instructions encoded by a frozen BERT-base-uncased~\cite{devlin2019bert} model and issued interactively by a human user: after each transport stage is completed, a new instruction specifying the next object and its goal is provided (e.g., \emph{``carry the laptop to the desk''}, then after completion, \emph{``move the trash bin to the corner''}). The robot must execute each instruction online without access to any privileged state information. Despite operating under realistic perceptual constraints, both VLA variants retain the long-horizon multi-object loco-manipulation capability of the teacher, suggesting that the behaviors learned by Humanoid Horizon are transferable to perception-driven settings and represent a meaningful step toward deployment on physical humanoid platforms.

\section{Experiments}

\subsection{Dataset and Evaluation Metrics}
\label{subsec:dataset}

Our experiments use the dataset from LHM-Humanoid~\cite{zhang2026lhmhumanoidlearningunifiedpolicy}, spanning four room categories---bedroom, living room, kitchen, and warehouse---with 350 diverse cluttered scene layouts and corresponding long-horizon loco-manipulation episodes. The scenes include 79 object categories, among which 25 are movable placement targets. In each episode, the agent must finish two consecutive fetch-carry-place cycles within one uninterrupted rollout (i.e., without resets). For generalization evaluation, we also use the 66 unseen tasks from LHM-Humanoid~\cite{zhang2026lhmhumanoidlearningunifiedpolicy}, which serve as an out-of-distribution benchmark with novel layouts, objects, and instructions.

We evaluate performance using \textbf{Succ1} and \textbf{Succ2} (success rates for each transport cycle), \textbf{SuccAll} (episodes where both cycles are completed), and \textbf{Dist~1}/\textbf{Dist~2} (final object-to-goal distances in meters). A cycle is considered successful if the object root lies within 0.2\,m of its target location. Better performance corresponds to higher success metrics and smaller distance errors.

\subsection{Baseline clarifications}
If parallel training is removed, our Humanoid Horizon effectively becomes Curriculum RL; in that case, dynamic starting and reward gating are no longer well-defined because both mechanisms are designed on top of the parallel-training framework. Hierarchical RL is implemented with five low-level controllers and a privileged FSM for oracle stage transitions: walk-to-target, reach-and-grasp, lift, carry with obstacle avoidance, and place-and-release. For HumanVLA~\cite{xu2024humanvla}, we decompose the 350 long-horizon tasks into 700 short-horizon subtasks to match its original protocol, since HumanVLA~\cite{xu2024humanvla} is built for single-object transport; reward design and training procedures are kept the same, and a privileged FSM is still used at test time to transition between consecutive transport stages. InterMimic~\cite{xu2025intermimic} is run according to its standard pipeline, i.e., AMP-motion imitation followed by policy rollout evaluation. TokenHSI~\cite{pan2025tokenhsi} uses the TokenHSI~\cite{pan2025tokenhsi} architecture while being trained on the same five skills as Hierarchical RL. The LHM-Humanoid~\cite{zhang2026lhmhumanoidlearningunifiedpolicy} baseline is likewise trained strictly under its original pipeline, and only its end-to-end student model is used for comparison. Except for the VLA table, all compared methods use identical observation and action interfaces, so differences primarily come from the learning strategies rather than mismatched I/O settings.

\subsection{Main results on 350 tasks}

\begin{table*}[htbp]
  \centering
  \caption{Results on 350 training tasks.}
  \label{tab:main_results}
  \begin{tabular}{lccccc}
    \toprule
    Method & Succ1 $\uparrow$ & Succ2 $\uparrow$ & SuccAll $\uparrow$ & Dist 1 (m) $\downarrow$ & Dist 2 (m) $\downarrow$ \\
    \midrule
    End-to-End RL          & 2.3\%  & 0.0\%  & 0.0\%  & 2.05 & 1.97 \\
    Curriculum RL          & 88.3\% & 48.5\% & 47.2\% & 0.26 & 1.01 \\
    Hierarchical RL        & 37.5\% & 25.0\% & 20.8\% & 0.78 & 1.13 \\
    HumanVLA~\cite{xu2024humanvla}  & 42.3\% & 36.7\% & 29.9\% & 0.65 & 0.97 \\
    InterMimic~\cite{xu2025intermimic} & 20.6\% & 9.8\%  & 7.5\%  & 1.32 & 1.82 \\
    TokenHSI~\cite{pan2025tokenhsi}   & 40.5\% & 31.5\% & 27.6\% & 0.58 & 1.07 \\
    LHM-Humanoid~\cite{zhang2026lhmhumanoidlearningunifiedpolicy}& 87.4\% & 72.6\% & 71.1\% & 0.28 & 0.50 \\
    Humanoid Horizon w/o DS   & 92.7\% & 73.5\% & 67.7\%  & 0.08 & 0.46 \\
    Humanoid Horizon w/o RG   & 83.9\% & 75.6\% & 69.6\% & 0.12 & 0.47 \\
    \textbf{Humanoid Horizon}& \textbf{96.6\%} & \textbf{81.3\%} & \textbf{80.4\%} & \textbf{0.06} & \textbf{0.38} \\
    \bottomrule
  \end{tabular}
\end{table*}

We report results on the 350 training tasks, 66 unseen tasks, and the VLA extension. DS and RG denote dynamic starting and reward gating, and three-seed standard deviations for our method and LHM-Humanoid are given in Appendix~\ref{app:seed_eval}. Table~\ref{tab:main_results} reveals clear failure patterns across baselines. End-to-End RL collapses in the long-horizon regime (SuccAll: 0.0\%), showing that sparse rewards do not propagate reliably through multi-stage, contact-rich execution. Curriculum RL exhibits easy-reward bias: despite high Succ1 (88.3\%), it under-learns transition behaviors and drops to 47.2\% SuccAll. Hierarchical RL, TokenHSI, and InterMimic also generalize poorly to cluttered rearrangement because narrow primitives, motion-derived controllers, or missing scene-specific supervision cause errors to accumulate across stages (SuccAll: 20.8\%, 27.6\%, and 7.5\%; Dist2 for InterMimic: 1.82\,m). HumanVLA further degrades to 29.9\% SuccAll, consistent with catastrophic forgetting caused by cumulative cross-stage distribution shift. LHM-Humanoid, the strongest prior baseline, still trails our method by about 9\% on Succ1/Succ2/SuccAll (87.4/72.6/71.1 vs.\ 96.6/81.3/80.4), highlighting the limits of a non-parallel design tuned mainly for two-object horizons.

Ablations confirm the complementary roles of DS and RG. Removing DS (\emph{w/o DS}) preserves strong first-stage performance (Succ1: 92.7\%) but makes downstream stages brittle because fixed proxy starts mismatch the evolving handoff states from preceding stages, reducing SuccAll to 67.7\%. Removing RG (\emph{w/o RG}) lowers both stability and full-horizon completion (SuccAll: 69.6\%) as previously placed objects are disturbed during later navigation. Combining Parallel Training, Dynamic Starting, and Reward Gating yields 96.6\%/81.3\%/80.4\% on Succ1/Succ2/SuccAll, establishing a new state of the art on this benchmark.

\subsection{Results on 66 unseen tasks}

\begin{table}[htbp]
  \centering
  \caption{Results on 66 unseen tasks.}
  \label{tab:unseen_results}
  \begin{tabular}{lccccc}
    \toprule
    Method & Succ1 $\uparrow$ & Succ2 $\uparrow$ & SuccAll $\uparrow$ & Dist 1 (m) $\downarrow$ & Dist 2 (m) $\downarrow$ \\
    \midrule
    End-to-End RL          & 1.4\%  & 0.0\%  & 0.0\%  & 2.37 & 2.28 \\
    Curriculum RL          & 75.5\% & 40.4\% & 39.5\% & 0.36 & 1.11 \\
    Hierarchical RL        & 31.6\% & 21.8\% & 17.3\% & 0.97 & 1.39 \\
    HumanVLA~\cite{xu2024humanvla}  & 36.3\% & 31.3\% & 25.1\% & 0.72 & 1.16 \\
    InterMimic~\cite{xu2025intermimic} & 17.2\% & 8.2\%  & 5.9\%  & 1.47 & 2.03 \\
    TokenHSI~\cite{pan2025tokenhsi}   & 33.9\% & 26.1\% & 22.9\% & 0.67 & 1.21 \\
    LHM-Humanoid~\cite{zhang2026lhmhumanoidlearningunifiedpolicy}    & 80.8\% & 64.9\% & 61.6\% & 0.37 & 0.54 \\
    Humanoid Horizon w/o DS   & 90.4\% & 69.8\% & 63.5\%  & 0.10 & 0.50 \\
    Humanoid Horizon w/o RG   & 89.1\% & 72.3\% & 66.4\% & 0.11 & 0.49 \\
    \textbf{Humanoid Horizon}& \textbf{97.1\%} & \textbf{79.3\%} & \textbf{78.0\%} & \textbf{0.05} & \textbf{0.47} \\
    \bottomrule
  \end{tabular}
\end{table}

Table~\ref{tab:unseen_results} evaluates generalization to 66 out-of-distribution scenes. The trend is consistent with the 350-task setting but with lower absolute values, indicating genuine generalization difficulty rather than memorization. Baselines still degrade substantially (e.g., End-to-End RL: SuccAll 0.0\%; Curriculum RL: 39.5\%; Hierarchical RL/TokenHSI: 17.3\%/22.9\%; HumanVLA: 25.1\%; InterMimic: 5.9\%). The strongest prior baseline, LHM-Humanoid~\cite{zhang2026lhmhumanoidlearningunifiedpolicy}, reaches 61.6\% SuccAll, while Humanoid Horizon achieves 97.1\%/79.3\%/78.0\% (Succ1/Succ2/SuccAll), outperforming it by about 16\% on SuccAll with only a marginal drop from seen tasks. Ablations remain consistent: removing Dynamic Starting (\emph{w/o DS}) reduces SuccAll to 63.5\%, while removing Reward Gating (\emph{w/o RG}) yields 89.1\%/72.3\%/66.4\% (Succ1/Succ2/SuccAll), still below the full model.

\subsection{VLA extension results}

\begin{table}[htbp]
  \centering
  \caption{VLA extension results (all methods distilled via the same DAgger~\cite{ross2011reduction} pipeline).}
  \label{tab:vla_results}
  \begin{tabular}{lccccc}
    \toprule
    Method & Succ1 $\uparrow$ & Succ2 $\uparrow$ & SuccAll $\uparrow$ & Dist 1 (m) $\downarrow$ & Dist 2 (m) $\downarrow$ \\
    \midrule
    End-to-End RL          & 1.3\%  & 0.0\%  & 0.0\%  & 2.50 & 2.49 \\
    Curriculum RL          & 71.5\% & 41.1\% & 40.7\% & 0.37 & 1.18 \\
    Hierarchical RL        & 29.3\% & 20.7\% & 16.5\% & 1.04 & 1.45 \\
    HumanVLA~\cite{xu2024humanvla}  & 34.2\% & 29.8\% & 23.5\% & 0.70 & 1.26 \\
    InterMimic~\cite{xu2025intermimic} & 16.3\% & 8.4\%  & 6.2\%  & 1.64 & 2.22 \\
    TokenHSI~\cite{pan2025tokenhsi}   & 32.0\% & 27.0\% & 23.7\% & 0.72 & 1.31 \\
    LHM-Humanoid~\cite{zhang2026lhmhumanoidlearningunifiedpolicy}    & 76.5\% & 65.4\% & 63.7\% & 0.42 & 0.56 \\
    Humanoid Horizon w/o DS   & 77.8\% & 58.4\% & 51.4\%  & 0.19 & 0.68 \\
    Humanoid Horizon w/o RG   & 74.0\% & 67.3\% & 58.6\% & 0.21 & 0.64 \\
    \textbf{Humanoid Horizon (RGB)}& 93.5\% & 76.8\% & \textbf{75.6\%} & \textbf{0.07} & \textbf{0.50} \\
    \textbf{Humanoid Horizon (Depth)}& \textbf{93.7\%} & \textbf{77.6\%} & 73.7\% & \textbf{0.07} & 0.52 \\
    \bottomrule
  \end{tabular}
\end{table}

Table~\ref{tab:vla_results} evaluates distillability rather than the absolute capability of VLA models. All methods are distilled with the same DAgger~\cite{ross2011reduction} pipeline into VLA students conditioned on egocentric observations and sequential human language instructions, where a new instruction is issued after each transport stage. Distillation degrades all methods because privileged state inputs are replaced by partial egocentric vision, but weaker teacher policies suffer larger drops, while Humanoid Horizon preserves a substantially larger fraction of its original performance. Our RGB and depth variants outperform all distilled baselines on most metrics, showing that the long-horizon behaviors learned through parallel training, dynamic starting, and reward gating transfer effectively to the egocentric VLA setting.

\subsection{Extension to more than two objects sequential transport}

\begin{table*}[htbp]
  \centering
  \caption{More than two objects sequential transport results. LHM-Humanoid$^\dagger$ denotes the variant trained with five teacher policies strictly following the original LHM-Humanoid~\cite{zhang2026lhmhumanoidlearningunifiedpolicy} pipeline.}
  \label{tab:five-object}
  \begin{tabular}{lcccccc}
    \toprule
    Method
    & Succ1 $\uparrow$
    & Succ2 $\uparrow$
    & Succ3 $\uparrow$
    & Succ4 $\uparrow$
    & Succ5 $\uparrow$
    & SuccAll $\uparrow$ \\
    \midrule
    End-to-End RL
    & 1.5\%  & 0.0\%  & 0.0\%  & 0.0\%  & 0.0\%  & 0.0\%  \\
    Curriculum RL
    & 91.2\% & 54.2\% & 37.5\% & 25.6\% & 9.3\%  & 3.2\%  \\
    Hierarchical RL
    & 38.6\% & 27.3\% & 16.6\% & 3.8\%  & 0.0\%  & 0.0\%  \\
    HumanVLA~\cite{xu2024humanvla}
    & 45.8\% & 40.5\% & 21.7\% & 5.6\%  & 1.2\%  & 0.1\%  \\
    InterMimic~\cite{xu2025intermimic}
    & 21.3\% & 10.2\% & 2.5\%  & 0.4\%  & 0.0\%  & 0.0\%  \\
    TokenHSI~\cite{pan2025tokenhsi}
    & 41.7\% & 32.7\% & 18.9\% & 4.9\%  & 1.1\%  & 0.1\%  \\
    LHM-Humanoid~\cite{zhang2026lhmhumanoidlearningunifiedpolicy}
    & 90.8\% & 76.1\% & 61.0\% & 38.5\% & 20.9\% & 18.1\% \\
    LHM-Humanoid$^\dagger$~\cite{zhang2026lhmhumanoidlearningunifiedpolicy}
    & 92.5\% & 78.8\% & 68.1\% & 48.6\% & 30.2\% & 25.8\% \\
    Humanoid Horizon w/o DS
    & 90.8\% & 77.4\% & 69.2\% & 61.4\% & 53.8\% & 49.8\% \\
    Humanoid Horizon w/o RG
    & 87.3\% & 76.9\% & 65.4\% & 54.8\% & 43.6\% & 38.9\% \\
    \textbf{Humanoid Horizon}
    & \textbf{96.5\%} & \textbf{85.2\%} & \textbf{82.2\%} & \textbf{75.4\%} & \textbf{66.5\%} & \textbf{60.2\%} \\
    \bottomrule
  \end{tabular}
\end{table*}

Table~\ref{tab:five-object} evaluates five-object sequential transport (five consecutive fetch-carry-place cycles without reset). Hierarchical RL, InterMimic, and TokenHSI are not retrained for longer horizons and are kept identical to their zero-shot setting, since their low-level primitives are learned independently of episode length; their results therefore reflect architectural limits. End-to-End RL is retrained under the five-object setting but still collapses (Succ1: 1.5\%). Curriculum RL is also retrained on five-object episodes, yet easy-reward bias remains and SuccAll drops to 3.2\%. LHM-Humanoid$^\dagger$, trained strictly with the original LHM pipeline using five teacher policies, improves over the two-teacher variant on later stages (Succ3--Succ5: 68.1\%/48.6\%/30.2\% vs.\ 61.0\%/38.5\%/20.9\%), but still degrades with horizon length.

Humanoid Horizon declines more slowly. Success falls from 82.2\% at stage 3 to 66.5\% at stage 5, with SuccAll at 60.2\%. Relative to the baselines just described, the drop is graceful, and the same policy stays ahead at every stage. As in the two-object setting, both added mechanisms matter: SuccAll falls to 49.8\% without Dynamic Starting and to 38.9\% without Reward Gating.

\section{Failure Analysis}

Humanoid Horizon exhibits two recurring failure modes. First, the Succ1-to-Succ2 gap is larger than later per-stage drops. The reason is an inherent task property: the first cycle starts from a fixed canonical pose on open floor, while later cycles begin from arbitrary post-task body configurations in clutter where surrounding objects become nearby obstacles, making reorientation and obstacle-aware navigation more difficult.

Second, frozen-action failures occur when the humanoid reaches the target but fails to complete a stable grasp. A likely contributor is uniform scene allocation during training: harder clutter layouts and object configurations require more exploration, so equal per-scene training budget can leave difficult grasp cases under-optimized. These challenges are not unique to our method and are also reflected in baseline degradation under longer horizons.

\section{Conclusion and Future Work}

We presented Humanoid Horizon, a unified framework for stable and scalable long-horizon whole-body humanoid loco-manipulation built on three tightly coupled mechanisms: Parallel Training, which maintains simultaneous optimization pressure across stages to mitigate easy-reward bias and catastrophic forgetting; Dynamic Starting, which trains downstream stages from realistic handoff-state distributions rather than fixed proxies; and Reward Gating, which preserves previously completed placements during later-stage transport. Across both in-distribution and out-of-distribution evaluation, this design delivers state-of-the-art performance on seen and unseen tasks, with strong gains over prior baselines on per-stage success and full-episode completion. Beyond the two-object setting, the same policy extends to five-object sequential transport without any change of architecture. Success declines as the horizon lengthens, but the degradation is graceful relative to the sharp drop of competing methods.

For future work, we plan to address the remaining failure modes (the Succ1-to-Succ2 navigation gap and frozen-action grasping failures) through improved path planning and difficulty-aware training, and to scale beyond five-object horizons toward real-world deployment on physical humanoid platforms.

\begin{ack}
This work was partially supported by the ERC Advanced project ETALK (UKRI Funded), the Horizon/UKRI project PRIMI and the US AFOSR project CASPER++.
\end{ack}

\bibliographystyle{unsrt}
{
  \small
  \bibliography{main,camera_ready_refs}

@inproceedings{devlin2019bert,
  title={Bert: Pre-training of deep bidirectional transformers for language understanding},
  author={Devlin, Jacob and Chang, Ming-Wei and Lee, Kenton and Toutanova, Kristina},
  booktitle={Proceedings of the 2019 conference of the North American chapter of the association for computational linguistics: human language technologies, volume 1 (long and short papers)},
  pages={4171--4186},
  year={2019}
}

@inproceedings{florensa2017reverse,
  title={Reverse curriculum generation for reinforcement learning},
  author={Florensa, Carlos and Held, David and Wulfmeier, Markus and Zhang, Michael and Abbeel, Pieter},
  booktitle={Conference on robot learning},
  pages={482--495},
  year={2017},
  organization={PMLR}
}

@article{eysenbach2017leave,
  title={Leave no trace: Learning to reset for safe and autonomous reinforcement learning},
  author={Eysenbach, Benjamin and Gu, Shixiang and Ibarz, Julian and Levine, Sergey},
  journal={arXiv preprint arXiv:1711.06782},
  year={2017}
}

@article{dai2021automatic,
  title={Automatic curricula via expert demonstrations},
  author={Dai, Siyu and Hofmann, Andreas and Williams, Brian},
  journal={arXiv preprint arXiv:2106.09159},
  year={2021}
}

@misc{zhang2026lhmhumanoidlearningunifiedpolicy,
      title={LHM-Humanoid: Learning a Unified Policy for Long-Horizon Humanoid Whole-Body Loco-Manipulation in Diverse Messy Environments}, 
      author={Haozhuo Zhang and Jingkai Sun and Michele Caprio and Jian Tang and Shanghang Zhang and Qiang Zhang and Wei Pan},
      year={2026},
      eprint={2508.16943},
      archivePrefix={arXiv},
      primaryClass={cs.RO},
      url={https://arxiv.org/abs/2508.16943}, 
}

@article{peng2021amp,
  title={Amp: Adversarial motion priors for stylized physics-based character control},
  author={Peng, Xue Bin and Ma, Ze and Abbeel, Pieter and Levine, Sergey and Kanazawa, Angjoo},
  journal={ACM Transactions on Graphics (ToG)},
  volume={40},
  number={4},
  pages={1--20},
  year={2021},
  publisher={ACM New York, NY, USA}
}

@inproceedings{tessler2023calm,
  title={Calm: Conditional adversarial latent models for directable virtual characters},
  author={Tessler, Chen and Kasten, Yoni and Guo, Yunrong and Mannor, Shie and Chechik, Gal and Peng, Xue Bin},
  booktitle={ACM SIGGRAPH 2023 Conference Proceedings},
  pages={1--9},
  year={2023}
}

@inproceedings{luo2023perpetual,
  title={Perpetual humanoid control for real-time simulated avatars},
  author={Luo, Zhengyi and Cao, Jinkun and Kitani, Kris and Xu, Weipeng and others},
  booktitle={Proceedings of the IEEE/CVF International Conference on Computer Vision},
  pages={10895--10904},
  year={2023}
}

@article{zhu2023neural,
  title={Neural categorical priors for physics-based character control},
  author={Zhu, Qingxu and Zhang, He and Lan, Mengting and Han, Lei},
  journal={ACM Transactions on Graphics (TOG)},
  volume={42},
  number={6},
  pages={1--16},
  year={2023},
  publisher={ACM New York, NY, USA}
}

@article{starke2019neural,
  title={Neural state machine for character-scene interactions},
  author={Starke, Sebastian and Zhang, He and Komura, Taku and Saito, Jun},
  journal={ACM Transactions on Graphics},
  volume={38},
  number={6},
  pages={178},
  year={2019},
  publisher={ACM}
}

@inproceedings{hassan2023synthesizing,
  title={Synthesizing physical character-scene interactions},
  author={Hassan, Mohamed and Guo, Yunrong and Wang, Tingwu and Black, Michael and Fidler, Sanja and Peng, Xue Bin},
  booktitle={ACM SIGGRAPH 2023 Conference Proceedings},
  pages={1--9},
  year={2023}
}

@article{xiao2023unified,
  title={Unified human-scene interaction via prompted chain-of-contacts},
  author={Xiao, Zeqi and Wang, Tai and Wang, Jingbo and Cao, Jinkun and Zhang, Wenwei and Dai, Bo and Lin, Dahua and Pang, Jiangmiao},
  journal={arXiv preprint arXiv:2309.07918},
  year={2023}
}

@inproceedings{pan2024synthesizing,
  title={Synthesizing physically plausible human motions in 3d scenes},
  author={Pan, Liang and Wang, Jingbo and Huang, Buzhen and Zhang, Junyu and Wang, Haofan and Tang, Xu and Wang, Yangang},
  booktitle={2024 International Conference on 3D Vision (3DV)},
  pages={1498--1507},
  year={2024},
  organization={IEEE}
}

@article{wang2023physhoi,
  title={Physhoi: Physics-based imitation of dynamic human-object interaction},
  author={Wang, Yinhuai and Lin, Jing and Zeng, Ailing and Luo, Zhengyi and Zhang, Jian and Zhang, Lei},
  journal={arXiv preprint arXiv:2312.04393},
  year={2023}
}

@article{xie2023hierarchical,
  title={Hierarchical planning and control for box loco-manipulation},
  author={Xie, Zhaoming and Tseng, Jonathan and Starke, Sebastian and van de Panne, Michiel and Liu, C Karen},
  journal={Proceedings of the ACM on Computer Graphics and Interactive Techniques},
  volume={6},
  number={3},
  pages={1--18},
  year={2023},
  publisher={ACM New York, NY, USA}
}

@article{xu2024humanvla,
  title={Humanvla: Towards vision-language directed object rearrangement by physical humanoid},
  author={Xu, Xinyu and Zhang, Yizheng and Li, Yong-Lu and Han, Lei and Lu, Cewu},
  journal={Advances in Neural Information Processing Systems},
  volume={37},
  pages={18633--18659},
  year={2024}
}

@article{wang2024skillmimic,
  title={Skillmimic: Learning reusable basketball skills from demonstrations},
  author={Wang, Yinhuai and Zhao, Qihan and Yu, Runyi and Zeng, Ailing and Lin, Jing and Luo, Zhengyi and Tsui, Hok Wai and Yu, Jiwen and Li, Xiu and Chen, Qifeng and others},
  journal={arXiv e-prints},
  pages={arXiv--2408},
  year={2024}
}

@inproceedings{xu2025intermimic,
  title={Intermimic: Towards universal whole-body control for physics-based human-object interactions},
  author={Xu, Sirui and Ling, Hung Yu and Wang, Yu-Xiong and Gui, Liang-Yan},
  booktitle={Proceedings of the Computer Vision and Pattern Recognition Conference},
  pages={12266--12277},
  year={2025}
}

@inproceedings{pan2025tokenhsi,
  title={Tokenhsi: Unified synthesis of physical human-scene interactions through task tokenization},
  author={Pan, Liang and Yang, Zeshi and Dou, Zhiyang and Wang, Wenjia and Huang, Buzhen and Dai, Bo and Komura, Taku and Wang, Jingbo},
  booktitle={Proceedings of the Computer Vision and Pattern Recognition Conference},
  pages={5379--5391},
  year={2025}
}

@article{ding2025humanoid,
  title={Humanoid-vla: Towards universal humanoid control with visual integration},
  author={Ding, Pengxiang and Ma, Jianfei and Tong, Xinyang and Zou, Binghong and Luo, Xinxin and Fan, Yiguo and Wang, Ting and Lu, Hongchao and Mo, Panzhong and Liu, Jinxin and others},
  journal={arXiv preprint arXiv:2502.14795},
  year={2025}
}

@inproceedings{ross2011reduction,
  title={A reduction of imitation learning and structured prediction to no-regret online learning},
  author={Ross, St{\'e}phane and Gordon, Geoffrey and Bagnell, Drew},
  booktitle={Proceedings of the fourteenth international conference on artificial intelligence and statistics},
  pages={627--635},
  year={2011},
  organization={JMLR Workshop and Conference Proceedings}
}

@inproceedings{jiang2024scaling,
  title={Scaling up dynamic human-scene interaction modeling},
  author={Jiang, Nan and Zhang, Zhiyuan and Li, Hongjie and Ma, Xiaoxuan and Wang, Zan and Chen, Yixin and Liu, Tengyu and Zhu, Yixin and Huang, Siyuan},
  booktitle={Proceedings of the IEEE/CVF Conference on Computer Vision and Pattern Recognition},
  pages={1737--1747},
  year={2024}
}

@article{peng2018deepmimic,
  title={Deepmimic: Example-guided deep reinforcement learning of physics-based character skills},
  author={Peng, Xue Bin and Abbeel, Pieter and Levine, Sergey and Van de Panne, Michiel},
  journal={ACM Transactions On Graphics (TOG)},
  volume={37},
  number={4},
  pages={1--14},
  year={2018},
  publisher={ACM New York, NY, USA}
}

@article{yenamandra2023homerobot,
  title={Homerobot: Open-vocabulary mobile manipulation},
  author={Yenamandra, Sriram and Ramachandran, Arun and Yadav, Karmesh and Wang, Austin and Khanna, Mukul and Gervet, Theophile and Yang, Tsung-Yen and Jain, Vidhi and Clegg, Alexander William and Turner, John and others},
  journal={arXiv preprint arXiv:2306.11565},
  year={2023}
}

@inproceedings{weihs2021visual,
  title={Visual room rearrangement},
  author={Weihs, Luca and Deitke, Matt and Kembhavi, Aniruddha and Mottaghi, Roozbeh},
  booktitle={Proceedings of the IEEE/CVF conference on computer vision and pattern recognition},
  pages={5922--5931},
  year={2021}
}

@article{wu2023tidybot,
  title={Tidybot: Personalized robot assistance with large language models},
  author={Wu, Jimmy and Antonova, Rika and Kan, Adam and Lepert, Marion and Zeng, Andy and Song, Shuran and Bohg, Jeannette and Rusinkiewicz, Szymon and Funkhouser, Thomas},
  journal={Autonomous Robots},
  volume={47},
  number={8},
  pages={1087--1102},
  year={2023},
  publisher={Springer}
}

@article{li2023object,
  title={Object motion guided human motion synthesis},
  author={Li, Jiaman and Wu, Jiajun and Liu, C Karen},
  journal={ACM Transactions on Graphics (TOG)},
  volume={42},
  number={6},
  pages={1--11},
  year={2023},
  publisher={ACM New York, NY, USA}
}

@misc{lin2025simgenhoiphysicallyrealisticwholebody,
      title={SimGenHOI: Physically Realistic Whole-Body Humanoid-Object Interaction via Generative Modeling and Reinforcement Learning}, 
      author={Yuhang Lin and Yijia Xie and Jiahong Xie and Yuehao Huang and Ruoyu Wang and Jiajun Lv and Yukai Ma and Xingxing Zuo},
      year={2025},
      eprint={2508.14120},
      archivePrefix={arXiv},
      primaryClass={cs.RO},
      url={https://arxiv.org/abs/2508.14120}, 
}

@article{wang2023mimicplay,
  title={Mimicplay: Long-horizon imitation learning by watching human play},
  author={Wang, Chen and Fan, Linxi and Sun, Jiankai and Zhang, Ruohan and Fei-Fei, Li and Xu, Danfei and Zhu, Yuke and Anandkumar, Anima},
  journal={arXiv preprint arXiv:2302.12422},
  year={2023}
}

@inproceedings{lin2024hierarchical,
  title={Hierarchical human-to-robot imitation learning for long-horizon tasks via cross-domain skill alignment},
  author={Lin, Zhenyang and Chen, Yurou and Liu, Zhiyong},
  booktitle={2024 IEEE International Conference on Robotics and Automation (ICRA)},
  pages={2783--2790},
  year={2024},
  organization={IEEE}
}

@article{zhao2024tac,
  title={Tac-Man: Tactile-informed prior-free manipulation of articulated objects},
  author={Zhao, Zihang and Li, Yuyang and Li, Wanlin and Qi, Zhenghao and Ruan, Lecheng and Zhu, Yixin and Althoefer, Kaspar},
  journal={IEEE Transactions on Robotics},
  volume={41},
  pages={538--557},
  year={2024},
  publisher={IEEE}
}

@inproceedings{jiao2021efficient,
  title={Efficient task planning for mobile manipulation: a virtual kinematic chain perspective},
  author={Jiao, Ziyuan and Zhang, Zeyu and Wang, Weiqi and Han, David and Zhu, Song-Chun and Zhu, Yixin and Liu, Hangxin},
  booktitle={2021 IEEE/RSJ International Conference on Intelligent Robots and Systems (IROS)},
  pages={8288--8294},
  year={2021},
  organization={IEEE}
}

@inproceedings{zhi2025closed,
  title={Closed-loop open-vocabulary mobile manipulation with gpt-4v},
  author={Zhi, Peiyuan and Zhang, Zhiyuan and Zhao, Yu and Han, Muzhi and Zhang, Zeyu and Li, Zhitian and Jiao, Ziyuan and Jia, Baoxiong and Huang, Siyuan},
  booktitle={2025 IEEE International Conference on Robotics and Automation (ICRA)},
  pages={4761--4767},
  year={2025},
  organization={IEEE}
}

@article{zhi2025learning,
  title={Learning unified force and position control for legged loco-manipulation},
  author={Zhi, Peiyuan and Li, Peiyang and Yin, Jianqin and Jia, Baoxiong and Huang, Siyuan},
  journal={arXiv e-prints},
  pages={arXiv--2505},
  year={2025}
}

@article{bu2025univla,
  title={Univla: Learning to act anywhere with task-centric latent actions},
  author={Bu, Qingwen and Yang, Yanting and Cai, Jisong and Gao, Shenyuan and Ren, Guanghui and Yao, Maoqing and Luo, Ping and Li, Hongyang},
  journal={arXiv preprint arXiv:2505.06111},
  year={2025}
}

@inproceedings{li2025clone,
  title={Clone: Closed-loop whole-body humanoid teleoperation for long-horizon tasks},
  author={Li, Yixuan and Lin, Yutang and Cui, Jieming and Liu, Tengyu and Liang, Wei and Zhu, Yixin and Huang, Siyuan},
  booktitle={9th Annual Conference on Robot Learning},
  year={2025}
}

@article{fan2025diffusion,
  title={Diffusion trajectory-guided policy for long-horizon robot manipulation},
  author={Fan, Shichao and Yang, Quantao and Liu, Yajie and Wu, Kun and Che, Zhengping and Liu, Qingjie and Wan, Min},
  journal={IEEE Robotics and Automation Letters},
  year={2025},
  publisher={IEEE}
}

@article{feng2025reflective,
  title={Reflective planning: Vision-language models for multi-stage long-horizon robotic manipulation},
  author={Feng, Yunhai and Han, Jiaming and Yang, Zhuoran and Yue, Xiangyu and Levine, Sergey and Luo, Jianlan},
  journal={arXiv preprint arXiv:2502.16707},
  year={2025}
}

@article{zhang2024navid,
  title={Navid: Video-based vlm plans the next step for vision-and-language navigation},
  author={Zhang, Jiazhao and Wang, Kunyu and Xu, Rongtao and Zhou, Gengze and Hong, Yicong and Fang, Xiaomeng and Wu, Qi and Zhang, Zhizheng and Wang, He},
  journal={arXiv preprint arXiv:2402.15852},
  year={2024}
}

@inproceedings{shridhar2023perceiver,
  title={Perceiver-actor: A multi-task transformer for robotic manipulation},
  author={Shridhar, Mohit and Manuelli, Lucas and Fox, Dieter},
  booktitle={Conference on Robot Learning},
  pages={785--799},
  year={2023},
  organization={PMLR}
}

@inproceedings{borja2022affordance,
  title={Affordance learning from play for sample-efficient policy learning},
  author={Borja-Diaz, Jessica and Mees, Oier and Kalweit, Gabriel and Hermann, Lukas and Boedecker, Joschka and Burgard, Wolfram},
  booktitle={2022 International Conference on Robotics and Automation (ICRA)},
  pages={6372--6378},
  year={2022},
  organization={IEEE}
}

@article{wang2025lola,
  title={LoLA: Long Horizon Latent Action Learning for General Robot Manipulation},
  author={Wang, Xiaofan and Gao, Xingyu and Fu, Jianlong and Li, Zuolei and Fortier, Dean and Mullins, Galen and Kolobov, Andrey and Guo, Baining},
  journal={arXiv preprint arXiv:2512.20166},
  year={2025}
}

@article{he2024omnih2o,
  title={Omnih2o: Universal and dexterous human-to-humanoid whole-body teleoperation and learning},
  author={He, Tairan and Luo, Zhengyi and He, Xialin and Xiao, Wenli and Zhang, Chong and Zhang, Weinan and Kitani, Kris and Liu, Changliu and Shi, Guanya},
  journal={arXiv preprint arXiv:2406.08858},
  year={2024}
}

@article{cheng2024expressive,
  title={Expressive whole-body control for humanoid robots},
  author={Cheng, Xuxin and Ji, Yandong and Chen, Junming and Yang, Ruihan and Yang, Ge and Wang, Xiaolong},
  journal={arXiv preprint arXiv:2402.16796},
  year={2024}
}

@article{ji2024exbody2,
  title={Exbody2: Advanced expressive humanoid whole-body control},
  author={Ji, Mazeyu and Peng, Xuanbin and Liu, Fangchen and Li, Jialong and Yang, Ge and Cheng, Xuxin and Wang, Xiaolong},
  journal={arXiv preprint arXiv:2412.13196},
  year={2024}
}

@article{wisth2022vilens,
  title={Vilens: Visual, inertial, lidar, and leg odometry for all-terrain legged robots},
  author={Wisth, David and Camurri, Marco and Fallon, Maurice},
  journal={IEEE Transactions on Robotics},
  volume={39},
  number={1},
  pages={309--326},
  year={2022},
  publisher={IEEE}
}

@article{ou2024leg,
  title={Leg-kilo: Robust kinematic-inertial-lidar odometry for dynamic legged robots},
  author={Ou, Guangjun and Li, Dong and Li, Hanmin},
  journal={IEEE Robotics and Automation Letters},
  volume={9},
  number={10},
  pages={8194--8201},
  year={2024},
  publisher={IEEE}
}

@article{allshire2025visual,
  title={Visual imitation enables contextual humanoid control},
  author={Allshire, Arthur and Choi, Hongsuk and Zhang, Junyi and McAllister, David and Zhang, Anthony and Kim, Chung Min and Darrell, Trevor and Abbeel, Pieter and Malik, Jitendra and Kanazawa, Angjoo},
  journal={arXiv preprint arXiv:2505.03729},
  year={2025}
}

@article{lin2026lessmimic,
  title={LessMimic: Long-Horizon Humanoid Interaction with Unified Distance Field Representations},
  author={Lin, Yutang and Cui, Jieming and Li, Yixuan and Jia, Baoxiong and Zhu, Yixin and Huang, Siyuan},
  journal={arXiv preprint arXiv:2602.21723},
  year={2026}
}
}

\appendix

\section{Implementation Details}
\label{app:implementation}

\subsection{Humanoid Embodiment}
\label{app:humanoid}
The humanoid agent has 15 rigid bodies: pelvis, torso, head, right/left upper arm, right/left lower arm, right/left hand, right/left thigh, right/left shin, and right/left foot. The action space consists of 28 degrees of freedom (DoF) controlled via PD position targets. The network outputs $\mathbf{a} \in [-1, 1]^{28}$, which is mapped to PD targets through an affine transformation:
\begin{equation}
q_{\text{target}} = \mathbf{a} \odot \mathbf{s} + \mathbf{o},
\end{equation}
where $\mathbf{s}$ (scale) and $\mathbf{o}$ (offset) are per-DoF constants derived from the joint limits. For revolute joints (right/left elbow and knee, 4 DoFs), the range is first shrunk by a factor of 0.7, then $s_j = 0.5 \times (\bar{q}^{+}_j - \bar{q}^{-}_j)$ and $o_j = 0.5 \times (\bar{q}^{+}_j + \bar{q}^{-}_j)$ where $\bar{q}^{\pm}$ denote the shrunk limits. For spherical joints (8 joints $\times$ 3 axes = 24 DoFs), the per-axis limit is set to $\min(1.2 \times \max|\theta_{\text{limit}}|, \pi)$ symmetrically, then $s_j$ and $o_j$ are computed analogously. The final torques are computed by the PD controller:
\begin{equation}
\tau = k_p \cdot (q_{\text{target}} - q_{\text{current}}) - k_d \cdot \dot{q}_{\text{current}},
\end{equation}
where $k_p$ and $k_d$ are the per-joint stiffness and damping gains. The control frequency is 30\,Hz ($\texttt{control\_freq\_inv}=2$ at 60\,Hz simulation).

\subsection{Observation and Action Spaces}
\label{app:obs_action}

\paragraph{Teacher (RL) observation.} The teacher policy receives a privileged state vector $\mathbf{o}_{\text{priv}} \in \mathbb{R}^{250}$ composed of: (i) body proprioception for all 15 rigid bodies (position [3], rotation in 6D representation [6], linear velocity [3], angular velocity [3]), totaling $15 \times 15 = 225$ dimensions (minus 2 for the root height redundancy); (ii) current target object position and orientation [9]; (iii) current object pose and velocity [15]; and (iv) navigation guidance waypoint [3]. In addition, a Basis Point Set (BPS) representation $\mathbf{b} \in \mathbb{R}^{200 \times 3}$ encodes the object point cloud from 200 basis points. The BPS is processed by a dedicated MLP $[600 \to 512 \to 512 \to 128]$ with ReLU activations. All components of $\mathbf{o}_{\text{priv}}$ and the BPS features are individually normalized via running mean and standard deviation.

\paragraph{AMP discriminator observation.} The adversarial motion prior (AMP) discriminator receives $\mathbf{o}_{\text{AMP}} \in \mathbb{R}^{1310}$, consisting of 10 consecutive reference motion observation frames ($\texttt{num\_ref\_obs\_frames}=10$), each of dimension 131. Each frame is composed of: root height [1], root rotation in 6D representation [6], local root linear velocity [3], local root angular velocity [3], joint rotations in 6D representation for all 12 joints [72] (8 spherical joints $\times$ 6D + 4 revolute joints $\times$ 6D), 28 DoF velocities [28], local positions of 5 key bodies (left hand, right hand, left foot, right foot, head) [15], and local object position [3], totaling $1+6+3+3+72+28+15+3=131$ per frame. The 5 key bodies are specified via $\texttt{amp\_body\_name}$.

\paragraph{VLA student observation.} The student is trained by DAgger to reproduce the teacher's actions from onboard sensing, and its observation is separate from $\mathbf{o}_{\text{priv}}$. It receives four signals. (i) Body proprioception $\mathbf{o}_{\text{body}} \in \mathbb{R}^{223}$, the per-body state of all 15 rigid bodies (position, 6D rotation, linear velocity, and angular velocity), with the two redundant root-height dimensions removed. (ii) An egocentric camera image $\mathbf{I} \in \mathbb{R}^{3 \times 256 \times 256}$ (RGB) or $\mathbb{R}^{1 \times 256 \times 256}$ (depth), captured from the head-mounted camera at position $[0.103, 0, 0.175]$ relative to the head link with horizontal FoV of $90^\circ$. (iii) A natural language instruction embedding $\mathbf{t} \in \mathbb{R}^{768}$, obtained from a frozen BERT-base-uncased~\cite{devlin2019bert} encoder. (iv) The previous action $\mathbf{a}_{t-1} \in \mathbb{R}^{28}$.

\paragraph{Action space.} All policies output $\mathbf{a} \in \mathbb{R}^{28}$, corresponding to normalized actions for the 28 humanoid DoFs. Actions are clipped to $[-1, 1]$ and then mapped to PD position targets via the per-DoF affine transformation $q_{\text{target}} = \mathbf{a} \odot \mathbf{s} + \mathbf{o}$ described in Sec.~\ref{app:humanoid}.

\subsection{Network Architectures}
\label{app:network}

\paragraph{Teacher (AMP) network.} The teacher uses a shared-backbone architecture with separate actor, critic, and discriminator heads:

\begin{itemize}
    \item \textbf{BPS encoder}: MLP $[600 \to 512 \to 512 \to 128]$ with ReLU.
    \item \textbf{Actor}: MLP $[\text{obs\_dim} \to 1024 \to 1024 \to 512 \to 28]$ with ReLU, where $\text{obs\_dim} = \dim(\mathbf{o}_{\text{priv}}) + 128$.
    \item \textbf{Critic}: MLP $[\text{obs\_dim} \to 1024 \to 1024 \to 512 \to 1]$ with ReLU.
    \item \textbf{Discriminator}: MLP $[1310 \to 1024 \to 1024 \to 512 \to 1]$ with ReLU; the final layer has no bias and uses $\text{Uniform}(-1, 1)$ weight initialization.
    \item \textbf{Action distribution}: Gaussian $\mathcal{N}(\mu, \sigma^2)$ with $\sigma$ as a learnable parameter (not conditioned on observation), initialized to $\exp(-2.9) \approx 0.055$.
\end{itemize}

\paragraph{VLA student network.} The VLA student network comprises:
\begin{itemize}
    \item \textbf{Image backbone}: EfficientNet-B0 (classifier head removed), producing a 1280-dimensional feature vector. Images are resized to $256 \times 256$ and normalized with mean and std of 0.5.
    \item \textbf{Image pre-MLP}: $[1280 \to 128]$.
    \item \textbf{Text pre-MLP}: $[768 \to 512 \to 128]$, where $768$ is the hidden dimension of the frozen BERT-base-uncased encoder.
    \item \textbf{Actor}: $[\dim(\mathbf{o}_{\text{body}}) + 128 + 128 + 28 \to 1024 \to 512 \to 28]$ with residual blocks, from body proprioception, the image embedding, the language embedding, and the previous action.
    \item The student outputs deterministic actions (no stochastic sampling).
\end{itemize}

\subsection{Reinforcement Learning Hyperparameters}
\label{app:rl_hyperparams}

Table~\ref{tab:rl_hyperparams} lists the key hyperparameters for teacher policy training.

\begin{table}[h]
  \centering
  \caption{Teacher (RL) training hyperparameters.}
  \label{tab:rl_hyperparams}
  \begin{tabular}{lc}
    \toprule
    Hyperparameter & Value \\
    \midrule
    PPO epochs per iteration & 4 \\
    Horizon length (rollout steps) & 32 \\
    Number of mini-batches & 8 \\
    Discount factor $\gamma$ & 0.98 \\
    GAE $\lambda$ & 0.95 \\
    PPO clip $\epsilon$ & 0.2 \\
    Optimizer & Adam \\
    Learning rate & $5 \times 10^{-5}$ \\
    Adam $(\beta_1, \beta_2)$ & $(0.9, 0.999)$ \\
    Weight decay & 0.0 \\
    Gradient norm clipping & 5.0 \\
    Auto mixed precision (AMP) & True \\
    Advantage normalization & True \\
    Action clip & 1.0 \\
    Observation clip & 5.0 \\
    AMP replay buffer size & 400{,}000 \\
    AMP demo fetch batch size & 512 \\
    AMP replay keep probability & 0.03 \\
    Parallel environments & 16{,}384 total, divided evenly across the $S$ stage streams \\
    Maximum episode length (single-object) & 300 steps \\
    Maximum episode length (multi-object) & 600 steps \\
    Maximum training epochs & 16{,}000 \\
    Simulation frequency & 60\,Hz \\
    Control frequency & 30\,Hz \\
    \bottomrule
  \end{tabular}
\end{table}

\subsection{Loss Functions and Weights}
\label{app:loss}

The total training loss comprises actor, critic, bound, and AMP discriminator losses:
\begin{equation}
  \mathcal{L}_{\text{total}} = w_a \mathcal{L}_{\text{actor}} + w_c \mathcal{L}_{\text{critic}} + w_b \mathcal{L}_{\text{bound}} + w_d \mathcal{L}_{\text{disc}},
\end{equation}
where:
\begin{itemize}
  \item $\mathcal{L}_{\text{actor}}$: PPO clipped surrogate objective.
  \item $\mathcal{L}_{\text{critic}}$: MSE loss on normalized value estimates.
  \item $\mathcal{L}_{\text{bound}}$: Action bound regularization $= \max(\mu - \text{clip}, 0)^2 + \min(\mu + \text{clip}, 0)^2$.
  \item $\mathcal{L}_{\text{disc}}$: AMP discriminator loss, composed of:
    \begin{itemize}
      \item Binary cross-entropy prediction loss (positive=demo, negative=policy).
      \item Logit regularization: $L_2$ norm of the final discriminator layer weights.
      \item Gradient penalty: $\|\nabla_{\mathbf{o}_{\text{pos}}} f(\mathbf{o}_{\text{pos}})\|^2$ where $f$ is the discriminator output.
      \item Weight decay on all discriminator parameters.
    \end{itemize}
\end{itemize}

The loss weights are listed in Table~\ref{tab:loss_weights}.

\begin{table}[h]
  \centering
  \caption{Loss weights.}
  \label{tab:loss_weights}
  \begin{tabular}{lc}
    \toprule
    Loss term & Weight \\
    \midrule
    Actor ($w_a$) & 1.0 \\
    Critic ($w_c$) & 5.0 \\
    Bound mu ($w_b$) & 10.0 \\
    Discriminator total ($w_d$) & 5.0 \\
    \quad Disc prediction & 1.0 \\
    \quad Disc logit regularization & 0.01 \\
    \quad Disc gradient penalty & 5.0 \\
    \quad Disc weight decay & 0.0001 \\
    \bottomrule
  \end{tabular}
\end{table}

\subsection{Reward Design}
\label{app:reward}

Each transport stage uses an 8-term shaped reward function. The total task reward for transporting object $o$ is:
\begin{equation}
  r_{\text{task}} = \sum_{k=1}^{8} w_k \cdot r_k,
\end{equation}
with individual terms and weights defined in Table~\ref{tab:reward_terms}.

\begin{table}[h]
  \centering
  \caption{Reward function components. All reward terms are bounded in $[0, 1]$. The thresholds $\delta_r = 0.5$\,m and $\delta_o = 0.3$\,m gate the approach and grasp rewards once the robot/object is sufficiently close. When $d_{\text{obj} \to \text{goal}} < \delta_o$, the approach, grasp, and height rewards are clamped to 1.0 to avoid interfering with placement.}
  \label{tab:reward_terms}
  \begin{tabular}{clcl}
    \toprule
    $w_k$ & Reward term & Name & Formula \\
    \midrule
    0.1 & Approach velocity & $r_{\text{vel}}$ & $\exp\!\bigl(-2(v^* - v_{\text{proj}})^2\bigr)$ \\
    0.1 & Approach position & $r_{\text{pos}}$ & $\exp\!\bigl(-0.5 \cdot d_{\text{robot} \to \text{obj}}\bigr)$ \\
    0.1 & Grasp proximity & $r_{\text{grasp}}$ & $\frac{1}{2}\sum_h \exp\!\bigl(-5 \cdot d_{h \to \text{obj}}\bigr)$ \\
    0.1 & Lift height & $r_{\text{height}}$ & $\frac{\min(z_{\text{obj}}, z_{\text{target}}) - z_{\text{init}}}{z_{\text{target}} - z_{\text{init}}}$ \\
    0.2 & Object transport vel. & $r_{\text{obj\_vel}}$ & $\exp\!\bigl(-2(v^*_{\text{obj}} - v_{\text{obj,proj}})^2\bigr)$ \\
    0.2 & Object waypoint dist. & $r_{\text{obj\_wp}}$ & $\exp\!\bigl(-1 \cdot d_{\text{obj} \to \text{waypoint}}\bigr)$ \\
    0.1 & Object goal proximity & $r_{\text{obj\_goal}}$ & $\exp\!\bigl(-5 \cdot d_{\text{obj} \to \text{goal}}\bigr)$ \\
    0.1 & Object goal rotation & $r_{\text{obj\_rot}}$ & $\exp\!\bigl(-2 \cdot \Delta\theta_{\text{obj} \to \text{goal}}\bigr)$ \\
    \bottomrule
  \end{tabular}
\end{table}

Here $v^* = v^*_{\text{obj}} = 1.5$\,m/s is the target navigation/transport speed, $v_{\text{proj}}$ is the projection of the robot/object velocity onto the direction toward the target, $d_{h \to \text{obj}}$ is the minimum distance from hand $h$ to the object point cloud, $z_{\text{target}} = \max(z_{\text{init}}, z_{\text{goal}}) + 0.1$ is the carry height (with a 0.1\,m lift margin), and $\Delta\theta$ is the geodesic rotation distance. For objects with both initial and goal heights below 0.1\,m (floor-to-floor transport), $r_{\text{height}}$ is set to 1.0.

The discriminator reward for style imitation is:
\begin{equation}
  r_{\text{disc}} = -\log\bigl(\max(1 - \sigma(f(\mathbf{o}_{\text{AMP}})),\; 0.01)\bigr),
\end{equation}
where $f$ is the AMP discriminator logit and $\sigma$ is the sigmoid function. The discriminator reward is clamped: $r_{\text{disc}} \leftarrow \text{clamp}(r_{\text{disc}}, 0, \max(r_{\text{task}}, 0.4))$. The final reward combines both:
\begin{equation}
  r = 1.0 \cdot r_{\text{disc}} + 1.0 \cdot r_{\text{task}}.
\end{equation}

\subsection{DAgger Distillation Hyperparameters}
\label{app:dagger}

Table~\ref{tab:dagger_hyperparams} lists the DAgger distillation hyperparameters.

\begin{table}[h]
  \centering
  \caption{DAgger distillation hyperparameters.}
  \label{tab:dagger_hyperparams}
  \begin{tabular}{lc}
    \toprule
    Hyperparameter & Value \\
    \midrule
    Initial mixing coefficient $\beta_0$ & 0.998 \\
    $\beta$ decay schedule & Exponential \\
    Number of step iterations per epoch & 1 \\
    Number of train iterations per epoch & 5 \\
    Replay buffer size & 100{,}000 \\
    Training batch size & 600 \\
    Test batch size & 200 \\
    Active rendering weight & 0.2 \\
    Optimizer & Adam \\
    Learning rate & $5 \times 10^{-4}$ \\
    Maximum training epochs & 30{,}000 \\
    Number of parallel environments & 558 \\
    Student loss & $\mathcal{L}_2 = \|\mathbf{a}_{\text{student}} - \mathbf{a}_{\text{teacher}}\|^2$ \\
    \bottomrule
  \end{tabular}
\end{table}

The DAgger training loop operates as follows: at each epoch, (1) the student executes an action in the environment with probability $1 - \beta$ (otherwise the teacher acts), (2) the teacher's action is recorded as the supervision target, and (3) the $(o_t, a^{\text{teacher}}_t)$ pairs are stored in the replay buffer. The student is then trained by sampling mini-batches from the buffer and minimizing the MSE loss. The mixing coefficient $\beta$ decays exponentially, gradually shifting from teacher-driven to student-driven rollouts.

\subsection{Simulation and Physics}
\label{app:simulation}

All experiments are conducted in NVIDIA IsaacGym with GPU-accelerated PhysX on 2$\times$ NVIDIA A100 (80\,GB) GPUs. Key physics settings:
\begin{itemize}
  \item Simulation timestep: $\Delta t = 1/60$\,s (60\,Hz) with 2 substeps.
  \item PhysX solver: TGS (Temporal Gauss-Seidel), 4 position iterations, 0 velocity iterations.
  \item Contact offset: 0.02\,m; max depenetration velocity: 10.0\,m/s.
  \item Ground plane: static friction~=~1.0, dynamic friction~=~1.0, restitution~=~0.0.
  \item Environment spacing: 5\,m between adjacent environments.
  \item Up axis: $z$.
\end{itemize}

\subsection{Termination Conditions}
\label{app:termination}

Episodes terminate under two conditions:
\begin{enumerate}
  \item \textbf{Fall detection}: A non-ignored body part (excluding hands, lower arms, and feet) has contact force $> 0.1$\,N and height $< 0.15$\,m simultaneously.
  \item \textbf{Maximum episode length}: 300 steps for single-object, 600 steps for two-object episodes.
\end{enumerate}

\section{Training Curves across Seeds}
\label{app:training_curves}

\begin{figure}[h]
  \centering
  \includegraphics[width=0.85\linewidth]{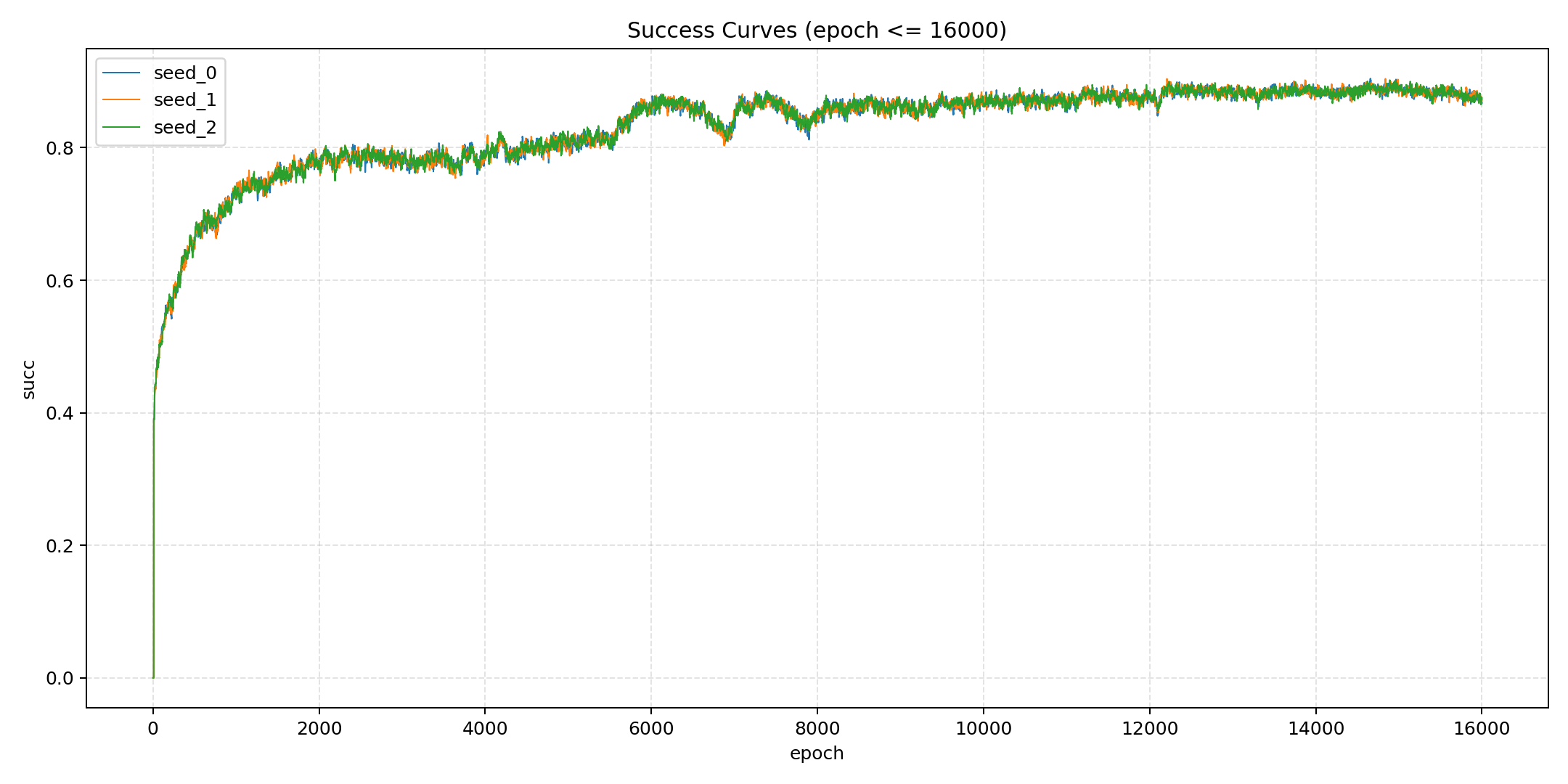}
  \caption{Training success curves across three random seeds (seed 0, 1, 2) over 16{,}000 epochs. Solid lines show per-seed success rates. All three seeds converge to similar final performance (SuccAll $\approx$ 87--89\%), demonstrating that Humanoid Horizon is robust to random initialization. The initial rapid increase (epochs 0--2{,}000) corresponds to the pretrained single-object checkpoint's adaptation to the parallel multi-object regime, followed by steady improvement through the interplay of dynamic starting and reward gating.}
  \label{fig:seed_curves}
\end{figure}

To verify the stability and reproducibility of Humanoid Horizon, we train the full pipeline with three independent random seeds (seed 0, 1, 2) on the 350-scene benchmark for 16{,}000 epochs each. Figure~\ref{fig:seed_curves} shows the training success curves for all three seeds.

Table~\ref{tab:seed_results} reports the final converged metrics (averaged over the last 100 epochs) for each seed, along with the mean and standard deviation across seeds.

\begin{table}[h]
  \centering
  \caption{Final training metrics across three random seeds (last 100 epochs). \texttt{success} refers to the SuccAll rate logged during training, and \texttt{success\_1} refers to the Succ1 rate.}
  \label{tab:seed_results}
  \begin{tabular}{lcc}
    \toprule
    Seed & success (SuccAll) & success\_1 (Succ1) \\
    \midrule
    Seed 0 & 87.7\% & 92.2\% \\
    Seed 1 & 87.3\% & 92.5\% \\
    Seed 2 & 87.8\% & 92.8\% \\
    \midrule
    Mean $\pm$ Std & 87.6 $\pm$ 0.3\% & 92.5 $\pm$ 0.3\% \\
    \bottomrule
  \end{tabular}
\end{table}

The low standard deviation ($\leq 0.3\%$) across seeds confirms that the Parallel Training Strategy, combined with Dynamic Starting and Reward Gating, yields stable convergence regardless of random initialization.

Two systematic differences exist between the training-time metrics reported here and the evaluation metrics in Table~\ref{tab:main_results} of the main paper. First, the training-time SuccAll (87.6\%) is \emph{higher} than the evaluation SuccAll (80.4\%). This is because Parallel Training flattens all $S$ transport stages into $S \times N$ independent concurrent streams; the logged SuccAll is therefore the average of per-stage success rates computed independently, rather than the product of sequential successes within a single uninterrupted episode. In evaluation, stages are executed sequentially---errors in earlier stages propagate and compound into later ones, naturally yielding a lower SuccAll. Second, the training-time Succ1 (92.5\%) is \emph{lower} than the evaluation Succ1 (96.6\%). This stems from a deliberately tighter success threshold used during training: as described in Appendix~\ref{app:termination}, the training-time success criterion requires the object to stay within $0.5 \times \epsilon_{\text{eval}} = 0.1$\,m, whereas evaluation counts a placement as successful if the object is within $\epsilon_{\text{eval}} = 0.2$\,m. This stricter training threshold encourages more precise object placement, which in turn improves evaluation robustness, but it also causes training-time per-stage success to be conservatively lower than the evaluation counterpart.

\section{Evaluation Seed Variance}
\label{app:seed_eval}

The tables in the main text report point estimates. Table~\ref{tab:seed_eval} gives the standard deviation of each success rate over the same three seeds, for Humanoid Horizon and LHM-Humanoid, in percentage points. The VLA block is the distilled student; RGB and depth are listed separately in Table~\ref{tab:vla_results}.

\begin{table}[h]
  \centering
  \caption{Three-seed standard deviations of success rates, in percentage points.}
  \label{tab:seed_eval}
  \begin{tabular}{lccc}
    \toprule
    & Succ1 & Succ2 & SuccAll \\
    \midrule
    \multicolumn{4}{l}{\textit{Seen tasks} (Table~\ref{tab:main_results})} \\
    LHM-Humanoid & 1.55 & 1.76 & 1.91 \\
    Humanoid Horizon & 0.94 & 0.13 & 0.94 \\
    \midrule
    \multicolumn{4}{l}{\textit{Unseen tasks} (Table~\ref{tab:unseen_results})} \\
    LHM-Humanoid & 3.27 & 3.78 & 3.78 \\
    Humanoid Horizon & 0.71 & 1.43 & 1.24 \\
    \midrule
    \multicolumn{4}{l}{\textit{VLA} (Table~\ref{tab:vla_results})} \\
    LHM-Humanoid & 5.23 & 5.73 & 4.36 \\
    Humanoid Horizon & 3.79 & 3.63 & 4.11 \\
    \bottomrule
  \end{tabular}

  \vspace{1.2em}

  \begin{tabular}{lcccccc}
    \toprule
    & Succ1 & Succ2 & Succ3 & Succ4 & Succ5 & SuccAll \\
    \midrule
    \multicolumn{7}{l}{\textit{Five-object transport} (Table~\ref{tab:five-object})} \\
    LHM-Humanoid & 1.66 & 2.15 & 1.48 & 3.48 & 3.16 & 2.76 \\
    Humanoid Horizon & 0.97 & 0.88 & 0.50 & 2.21 & 1.12 & 1.91 \\
    \bottomrule
  \end{tabular}
\end{table}

\section{Broader Impact}
\label{app:broader_impact}

This work advances autonomous humanoid manipulation in simulation, with potential positive impacts including assistive robotics (e.g., tidying rooms for elderly or mobility-impaired users) and warehouse automation. Potential risks include the dual-use concern that humanoid agents capable of autonomous object manipulation in cluttered environments could be repurposed for unintended applications. We mitigate this by releasing only the simulation-based framework and trained policies; real-world deployment would require additional safety mechanisms (e.g., force limits, human-in-the-loop supervision). The method does not use personal data, human subjects, or crowd-sourcing.

\clearpage
\section*{NeurIPS Paper Checklist}

\begin{enumerate}

\item {\bf Claims}
    \item[] Question: Do the main claims made in the abstract and introduction accurately reflect the paper's contributions and scope?
    \item[] Answer: \answerYes{}
    \item[] Justification: The claims in the abstract and introduction are supported by the experiments and ablations in Sec.~4.

\item {\bf Limitations}
    \item[] Question: Does the paper discuss the limitations of the work performed by the authors?
    \item[] Answer: \answerYes{}
    \item[] Justification: Limitations are discussed in Sec.~5 (Failure Analysis) and Sec.~6 (Future Work).

\item {\bf Theory assumptions and proofs}
    \item[] Question: For each theoretical result, does the paper provide the full set of assumptions and a complete (and correct) proof?
    \item[] Answer: \answerNA{}
    \item[] Justification: The paper does not include theoretical results.

    \item {\bf Experimental result reproducibility}
    \item[] Question: Does the paper fully disclose all the information needed to reproduce the main experimental results of the paper to the extent that it affects the main claims and/or conclusions of the paper (regardless of whether the code and data are provided or not)?
    \item[] Answer: \answerYes{}
    \item[] Justification: Full implementation details are provided in the appendix.

\item {\bf Open access to data and code}
    \item[] Question: Does the paper provide open access to the data and code, with sufficient instructions to faithfully reproduce the main experimental results, as described in supplemental material?
    \item[] Answer: \answerNo{}
    \item[] Justification: Dataset and code will be released upon acceptance.

\item {\bf Experimental setting/details}
    \item[] Question: Does the paper specify all the training and test details (e.g., data splits, hyperparameters, how they were chosen, type of optimizer) necessary to understand the results?
    \item[] Answer: \answerYes{}
    \item[] Justification: Data splits and metrics are described in Sec.~4.1, and all hyperparameters and training details are in the appendix.

\item {\bf Experiment statistical significance}
    \item[] Question: Does the paper report error bars suitably and correctly defined or other appropriate information about the statistical significance of the experiments?
    \item[] Answer: \answerYes{}
    \item[] Justification: Results are reported as mean $\pm$ standard deviation across three random seeds in Appendix~B.

\item {\bf Experiments compute resources}
    \item[] Question: For each experiment, does the paper provide sufficient information on the computer resources (type of compute workers, memory, time of execution) needed to reproduce the experiments?
    \item[] Answer: \answerYes{}
    \item[] Justification: Compute resources (GPUs, environments, epochs) are reported in the appendix.
    
\item {\bf Code of ethics}
    \item[] Question: Does the research conducted in the paper conform, in every respect, with the NeurIPS Code of Ethics \url{https://neurips.cc/public/EthicsGuidelines}?
    \item[] Answer: \answerYes{}
    \item[] Justification: The research is simulation-only and conforms to the NeurIPS Code of Ethics.

\item {\bf Broader impacts}
    \item[] Question: Does the paper discuss both potential positive societal impacts and negative societal impacts of the work performed?
    \item[] Answer: \answerYes{}
    \item[] Justification: Broader impacts are discussed in Appendix~D.
    
\item {\bf Safeguards}
    \item[] Question: Does the paper describe safeguards that have been put in place for responsible release of data or models that have a high risk for misuse (e.g., pre-trained language models, image generators, or scraped datasets)?
    \item[] Answer: \answerYes{}
    \item[] Justification: Released models are simulation-only and pose limited misuse risk; safeguards are briefly noted in Appendix~D.

\item {\bf Licenses for existing assets}
    \item[] Question: Are the creators or original owners of assets (e.g., code, data, models), used in the paper, properly credited and are the license and terms of use explicitly mentioned and properly respected?
    \item[] Answer: \answerYes{}
    \item[] Justification: All existing assets used are cited and used under their respective licenses.

\item {\bf New assets}
    \item[] Question: Are new assets introduced in the paper well documented and is the documentation provided alongside the assets?
    \item[] Answer: \answerYes{}
    \item[] Justification: Full details are in the appendix.

\item {\bf Crowdsourcing and research with human subjects}
    \item[] Question: For crowdsourcing experiments and research with human subjects, does the paper include the full text of instructions given to participants and screenshots, if applicable, as well as details about compensation (if any)? 
    \item[] Answer: \answerNA{}
    \item[] Justification: No crowdsourcing or human subjects are involved.

\item {\bf Institutional review board (IRB) approvals or equivalent for research with human subjects}
    \item[] Question: Does the paper describe potential risks incurred by study participants, whether such risks were disclosed to the subjects, and whether Institutional Review Board (IRB) approvals (or an equivalent approval/review based on the requirements of your country or institution) were obtained?
    \item[] Answer: \answerNA{}
    \item[] Justification: No human subjects research is involved.

\item {\bf Declaration of LLM usage}
    \item[] Question: Does the paper describe the usage of LLMs if it is an important, original, or non-standard component of the core methods in this research? Note that if the LLM is used only for writing, editing, or formatting purposes and does \emph{not} impact the core methodology, scientific rigor, or originality of the research, declaration is not required.
    \item[] Answer: \answerNA{}
    \item[] Justification: LLMs are not part of the core method.

\end{enumerate}

\end{document}